\pdfoutput=1
\documentclass[sigconf]{acmart}

\definecolor{gray}{RGB}{192,192,192}

\usepackage{smartdiagram}
\usepackage{tikz}
\usepackage{pgfplots}
\pgfplotsset{compat=1.17}
\usetikzlibrary{
    patterns,
}

\usepackage[linesnumbered,ruled,vlined]{algorithm2e}
\usepackage{booktabs}
\usepackage{bm}
\usepackage{caption}
\usepackage{subfigure}
\usepackage{color}
\usepackage{multirow}
\usepackage{footnote}
\usepackage{pifont}
\usepackage{cleveref}

\newtheorem{definition}{Definition}
\definecolor{MyPink}{RGB}{255,178,178}
\definecolor{MyBlue}{RGB}{178,178,255}
\PassOptionsToPackage{hyphens}{url}\usepackage{hyperref}
\hypersetup{colorlinks=true}
\usetikzlibrary{er,positioning}

\copyrightyear{2022} 
\acmYear{2022} 
\setcopyright{acmcopyright}\acmConference[CIKM '22]{Proceedings of the 31st ACM International Conference on Information and Knowledge Management}{October 17--21, 2022}{Atlanta, GA, USA}
\acmBooktitle{Proceedings of the 31st ACM International Conference on Information and Knowledge Management (CIKM '22), October 17--21, 2022, Atlanta, GA, USA}
\acmPrice{15.00}
\acmDOI{10.1145/3511808.3557264}
\acmISBN{978-1-4503-9236-5/22/10}

\title{Contrastive Knowledge Graph Error Detection}

\author{Qinggang Zhang}
\affiliation{%
  \institution{The Hong Kong Polytechnic University}
  \city{Hung Hom, Hong Kong SAR}
  }
\email{qinggangg.zhang@connect.polyu.hk}

\author{Junnan Dong}
\affiliation{%
  \institution{The Hong Kong Polytechnic University}
  \city{Hung Hom, Hong Kong SAR}
  }
\email{hanson.dong@connect.polyu.hk}

\author{Keyu Duan}
\affiliation{%
  \institution{The Hong Kong Polytechnic University}
  \city{Hung Hom, Hong Kong SAR}
  }
\email{k.duane9902@gmail.com}

\author{Xiao Huang}
\affiliation{%
 \institution{The Hong Kong Polytechnic University}
  \city{Hung Hom, Hong Kong SAR}
  }
\email{xiaohuang@comp.polyu.edu.hk}

\author{Yezi Liu}
\affiliation{%
  \institution{University of California Irvine}
  \city{Irvine}
  \state{CA}
  \country{USA}}
\email{yezil3@uci.edu}

\author{Linchuan Xu}
\affiliation{%
 \institution{The Hong Kong Polytechnic University}
  \city{Hung Hom, Hong Kong SAR}
  }
\email{linch.xu@polyu.edu.hk}

\renewcommand{\shortauthors}{Qinggang Zhang et al.}

\begin{document}

\begin{abstract}
Knowledge Graph (KG) errors introduce non-negligible noise, severely affecting KG-related downstream tasks. Detecting errors in KGs is challenging since the patterns of errors are unknown and diverse, while ground-truth labels are rare or even unavailable. A traditional solution is to construct logical rules to verify triples, but it is not generalizable since different KGs have distinct rules with domain knowledge involved. Recent studies focus on designing tailored detectors or ranking triples based on KG embedding loss. However, they all rely on negative samples for training, which are generated by randomly replacing the head or tail entity of existing triples. Such a negative sampling strategy is not enough for prototyping practical KG errors, e.g., ({\em Bruce\_Lee, place\_of\_birth, China}), in which the three elements are often relevant, although mismatched. We desire a more effective unsupervised learning mechanism tailored for KG error detection. To this end, we propose a novel framework - ContrAstive knowledge Graph Error Detection (CAGED). It introduces contrastive learning into KG learning and provides a novel way of modeling KG. Instead of following the traditional setting, i.e., considering entities as nodes and relations as semantic edges, CAGED augments a KG into different hyper-views, by regarding each relational triple as a node. After joint training with KG embedding and contrastive learning loss, 
CAGED assesses the trustworthiness of each triple based on two learning signals, i.e., the consistency of triple representations across multi-views and the self-consistency within the triple. Extensive experiments on three real-world KGs show that CAGED outperforms state-of-the-art methods in KG error detection. Our codes and datasets are available at {\small\url{https://github.com/Qing145/CAGED.git}}.
\end{abstract}




\maketitle

% \vspace{-2mm}
\section{Introduction} \label{sec:intro}
Knowledge graph (KG) error detection becomes increasingly essential as more KG-based systems have been being deployed, such as conversational agents, and recommender systems~\cite{Wang-etal19Knowledge,Huang-etal19Knowledge}. KGs reformulate a realistic assertion as a triple, i.e., \textit{(head entity, relation, tail entity)}, and could effectively organize information and knowledge in a structured and scalable way. 
Most KGs are extracted from web corpora by using heuristic algorithms. 
Amounts of noisy triples were inevitably introduced into KGs due to the noises in the original sources and the imperfect extraction algorithms~\citep{Bollacker-etal07Freebase, Lehmann-etal15DBpedia, Heindorf-etal16Vandalism,Mahdisoltani-etal15YAGO}. For example, a widely-used KG, NELL~\cite{Carlson-etal10Toward} has a precision of $74\%$, corresponding to around 0.6 million inaccurate triples. Existing KG driven studies ignore the significant impact of KG errors. They assume that all triples in KGs are correct and therefore leads to severe performance degradation in downstream tasks. To mitigate the roadblocks, there is an urgent need for the development of effective KG error detection algorithms.
% To mitigate the roadblocks, effective  methods are of vital importance. 

One of the key challenges in KG error detection is that the patterns of errors are unknown and diverse, while ground-truth labels are rare or even unavailable. In practice, trivial erroneous triples were often corrected during the construction of KGs~\cite{Paulheim17}. Therefore, labeled KG errors, i.e., the ground truth, are often not available, and remaining errors in KGs are nontrivial. Briefly, the taxonomy of previous works encompasses two branches: $(i)$ \textit{Rule-based} and $(ii)$ \textit{Embedding-based}. Namely, the former one conducts KG error detection based on rules~\cite{agrawal1993mining,guo2018knowledge}. They define KG errors as assertions that violate any one of the pre-defined rules~\cite{cheng2018rule,galarraga2013amie,tanon2017completeness}. But these methods are not generalizable since different KGs have distinct rules with domain knowledge involved.
More recently, several embedding-based KG error detection methods have been explored~\cite{Paulheim-Gangemi15Serving,jia2019triple,Ge-etal20KGClean,xu2022contrastive,Melo-Paulheim17Detection}. 
To perform unsupervised error detection, they employ a naive negative sampling to train models~\cite{jia2019triple}. Such negative samples have also been employed as synthetic labels to train detectors in supervised KG error detection methods~\cite{Ge-etal20KGClean,Melo-Paulheim17Detection}. 

However, negative samples employed by existing models bring an inherent limitation. Given a positive triple $(h,r,t)$, the corresponding negative samples are generated by randomly replacing the entities, i.e. $h$ or $t$. Such a negative sampling strategy is not enough for prototyping practical KG errors. For instance, given a true triple \textit{(Newton, Nationality, England)}, they may randomly generate \textit{(Newton, Nationality, Google)}. But real-world errors often have more complex patterns,  e.g., ({\em Bruce\_Lee, place\_of\_birth, China}), in which the three elements are mismatched but often relevant, since Lee was born in the Chinatown area of San Francisco. To this end, we desire a more appropriate way to enable effective detection.

As an alternative self-supervised learning technique~\cite{chen2020simple,henaff2020dataeff,hjelm2019learning}, contrastive learning has proven effective in many network analysis tasks~\cite{kipf2020contrastive,peng2020graph,hassani2020contrastive,zhu2020deep,you2020graph}, e.g. classification~\cite{ZengX21}, link prediction~\cite{WangMCW21}, and recommender systems~\cite{0013YYWC021}.
% Actually, contrastive learning can also be used as a promising solution to detect errors by learning and contrasting representation among different views, and simultaneously, address the aforementioned limitations.  
Previous study~\cite{tack2020csi, chen2021novelty} have already extended the power of contrastive learning for error detection in computer vision and demonstrates its superiority under various error detection scenarios.  Using data transformations, it first creates distinct views of the original data, and then learn the representations by maximizing agreement of each instance's representation among different views. By contrasting and learning  between the elaborate instance pairs, the model can capture informative features for distinguishing errors without manual labels. 
\begin{table}[!t]
  \caption{Major notations and definitions.}
  \label{tab:symbols}
\vspace{-0.2cm}
\begin{tabular}{c|c}
\hline
\textbf{Notations} & \textbf{Descriptions}  \\ 
\hline
$\mathcal{G}$ & a knowledge graph with errors\\
%$\mathcal{E}$   & Entity set\\
%$\mathcal{R}$   & Relation set\\
%$\mathcal{S}$   & Triple set\\
$(h, r, t)$ & a triple with head entity $h$, relation $r$, tail entity $t$\\
% \multirow{2}{*}{$\mathbf{E}$} & trainable embedding representation lookup table \\
&of all entities and all relation types\\
%$\mathbf{R}$    & {\red Trainable embedding representations of }\\
$(\mathbf{e}_{h}, \mathbf{e}_{r}, \mathbf{e}_{t})\!\!$  & embedding representations of $h$, $r$, and $t$\\
% $d$ & the dimension of embedding representations\\\
%$n$ & Total number of triples within a minbatch\\
%$m$ & Number of neighbors of a triple\\
$\mathcal{T}$   & view I of $\mathcal{G}$, i.e., a graph with triples as nodes\\
$\mathcal{T}'$   & view II of $\mathcal{G}$, i.e., a graph with triples as nodes\\
%${\bf q}_{i}$  & local embedding representation of $i^{\text {th}}$ triple\\
$\mathbf{x}_{i}$    & representation of $i^{\text {th}}$ triple learned from $\mathcal{T}$\\
${\bf z}_{i}$    & representation of $i^{\text {th}}$ triple learned from $\mathcal{T}'$\\
\hline
\end{tabular}
%}
\vspace{-6mm}
\end{table}

In spite of the wide applications of contrastive learning, we argue that it is nontrivial to explore it for KG error detection. Basically, there are two major challenges. First, defining views on KGs is not a trivial task, given that KGs and errors have unique data structure. Most existing graph contrastive learning frameworks rely on graph-level augmentations such as node dropping, edge perturbation, subgraph sampling or matrix diffusion~\cite{you2020graph,hassani2020contrastive} to create different views of graphs. However, these graph-level augmentation methods are not suitable for KG error detection. A pure KG is actually a set of triples, thus detecting errors on the KG is equal to identifying noisy triples. Existing graph contrastive learning frameworks, that focus on entity or graph-level contrasting, benefit little for learning rich and distinguishing representations at triple level. Additionally, existing graph augmentation methods are designed for network embedding, which could boost the robustness of the learned representations but at the price of damaging the underlying  distribution of potential errors.
Taking node dropping and edge perturbation as typical examples, they would change the structure of potential noisy triples or even introduce new erroneous triples, which makes the detection task more challenging. Second, existing graph encoders would not take anomalies or errors into consideration. As a result, potential erroneous triples may also contribute significantly during the message-passing process, which lead to suboptimal performance in error detection. Thus, a robust error-aware graph encoder is desired, aiming to alleviate the impact of errors and perform effective learning. 

To this end, we formally define the problem of KG error detection, and investigate two research questions. \ding{182} How to construct appropriate different views of a KG to conduct effective contrastive learning? \ding{183} How to perform error-aware KG encoding that is tailored for error detection? To answer these questions, we propose an effective framework named ContrAstive knowledge Graph Error Detection (CAGED). Our major contributions are listed as follows.
\begin{itemize}
\item We develop a framework - CAGED, which could perform effective error detection by integrating contrastive learning and KG embedding.
\item 
We design a new KG augmentation mechanism to enable appropriate contrastive learning in CAGED. It generates different views of a KG at the triple level and avoids undermining the underlying structure of potential errors.
\item We present an error-aware graph encoder - EaGNN, to model triple representations for each view. It adopts a tailored gated attention mechanism to block the information propagation from erroneous triples.
\item We conduct empirical studies on three real-world KGs. Experimental results show that CAGED outperforms state-of-the-art error detection algorithms.
\end{itemize}

\vspace{-2mm}
\section{Problem Statement} \label{sec:problem}
\noindent\textbf{Notations:} We use an uppercase bold alphabet (e.g., $\textbf{W}$) to denote a matrix and a lowercase bold alphabet (e.g., ${\textbf x}$) to represent a vector. The transpose of a matrix is denoted as ${\textbf W}^\top$. We use $\|{\textbf x}\|_2$ to represent the $\ell^2$ norm of a vector. The operation ${\textbf x} = [{\textbf h}; {\textbf r}; {\textbf t}]$ denotes concatenating column vectors ${\textbf h}$, ${\textbf r}$, and ${\textbf t}$ into a new column vector ${\textbf x}$. 
We list the major symbols in this paper in Table~\ref{tab:symbols}. 
Let $\mathcal{G}$ denote a KG that contains a set of triples. Each triple is composed of a head entity $h$, a relation $r$, and a tail entity $t$, represented as $(h,r,t)$. 
We formally define KG errors at the triple level. 
\begin{definition}\label{def:KG_errors}
\textbf{Knowledge Graph Errors.} Given a triple in a KG $(h,r,t)$, if there is a mismatch between head/tail entities and its relation $r$, then this triple $(h,r,t)$ is an error. E.g., (Bruce\_Lee, place\_of\_birth, China) and (Bill\_Gates, CEO, Google) are errors.
\end{definition}

\begin{figure*}
\centering
	\includegraphics[scale=0.285]{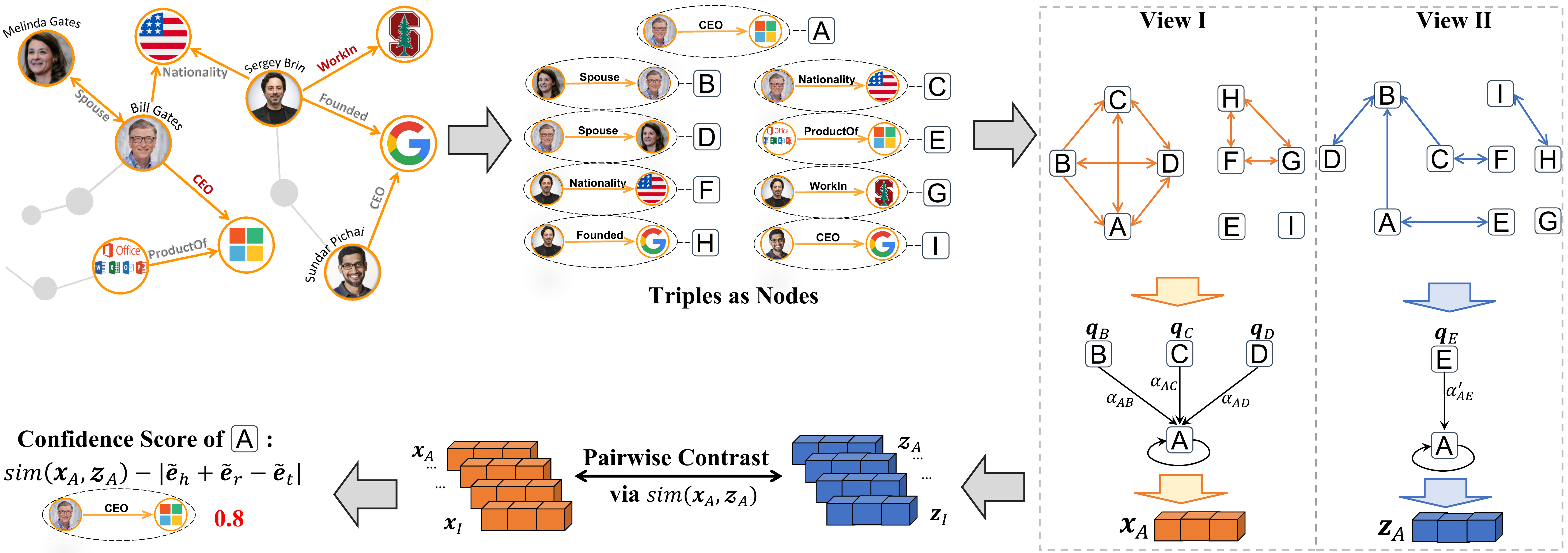}
	\vspace{-0.2cm}
	\caption{Two separate augmentation operators are applied to the original KG, generating two triple graphs as congruent views, i.e., View I and View II. After training, we estimate the confidence score by measuring the consistency of triple representations across multi-views, i.e., $\bm{x}_A$ and $\bm{z}_A$, and the self-consistency within the triple, i.e., $(\tilde{\bm{e}}_h, \tilde{\bm{e}}_r, \tilde{\bm{e}}_t)$.}
	\label{fig:caged}\vspace{-0.4cm}
\end{figure*}

In this paper, we employ a metric named confidence scores~\cite{xie2018does}, ranging from 0 to 1, to indicate the degree of triples being correct. False triples should have confidence scores close to 0. 
We formally define the problem of KG error detection as follows.

\vspace{2mm}
\noindent\fbox{\begin{minipage}{26.1em}
Given a KG $\mathcal{G}=\{(h,r,t)\}$, we aim to find an optimal rating model $p_{\omega}:(h,r,t) \rightarrow \mathbb{R}^{(0,1)}$ with our proposed training framework such that $p_\omega(h,r,t)$ accurately denotes the probability of triple $(h,r,t)$ being a defined error in Def~\ref{def:KG_errors}. The performance of the framework is evaluated by \textit{Precision} and \textit{Recall} of the triples with the top lowest confidence scores.
\end{minipage}}

\section{Methodology}\label{sec:method}
In traditional KG representation learning, a KG is usually modeled as a heterogeneous graph by centering the entities as nodes and regarding relations as semantic edges. However, with such modeling, it is notoriously hard to capture the complex correlation across triples. In this work, we propose a novel framework named ContrAstive knowledge Graph Error Detection (CAGED) to effectively detect errors on large-scale KGs. Its core idea is to transform the target KG into  different hyper-views, i.e. triple graphs, by regarding each relational triple as nodes, and then assess the trustworthiness of each triple based on the consistency between its representations learned from multiple views. The intuition behind this approach is that contrastive learning can learn a rich representation of key semantics in normal instances then the absence of such feature in anomalous triples can force their representations to lie far in the latent vector space, thus how well a triple's representations from different views can be pulled close to each other serves as a good learning signal for error detection. 

As illustrated in Figure~\ref{fig:caged}, our proposed CAGED consists of three components, i.e., KG augmentation, error-aware encoder, and joint confidence estimation.
First, we propose a novel KG augmentation method to generate two triple-level graphs by regarding each relational triple as nodes and treat them as two congruent views, i.e. View I and View II, of our contrastive framework.
Second, we design a tailored error-aware knowledge graph neural network (EaGNN) as the encoder to block the information propagation from potential errors during the representation learning phase.  Third, to learn rich and distinguishable across-view triple representations, we integrate translation-based KG embedding loss and contrastive learning loss to jointly optimize the model. After training, we estimate the confidence score of each triple based on two learning signals, i.e., the consistency of triple representations across multi-views and the self-consistency within the triple.

\subsection{Knowledge Graph Augmentation}\label{sec:KG_augmentation}
Data augmentation is the most crucial part in contrastive learning~\cite{kipf2020contrastive}. However, unlike image data in which views can be created by standard augmentations~\citep{zhu2020deep,hassani2020contrastive}, e.g., rotating, cropping, distorting, etc., defining views on KGs is not a trivial task due to its unique data structure. Most existing graph contrastive learning frameworks rely on graph-level augmentations such as node dropping, edge perturbation, subgraph sampling or matrix diffusion~\cite{you2020graph} to create different views of graphs, and achieve considerable performance improvements in some classical graph related downstream tasks, such as node classification, graph classification, and link prediction.  
 
 However, these graph-level augmentation methods are not suitable for KG error detection. A pure KG is actually a set of triples, thus detecting errors on the KG is equal to identifying noisy triples. Existing graph contrastive learning framework, that focus on entity or graph-level contrasting, benefit little for learning rich and distinguishing representations at triple level.
% Additionally, existing graph augmentation methods are designed for network embedding, which could boost the robustness of the learned representations and make errors less obvious.
 Additionally, existing graph augmentation methods are designed for network embedding, which could boost the robustness of the learned representations but at the price of damaging the underlying structure of potential errors. Specifically, existing augmentation methods such as node dropping, edge perturbation would change the distribution of potential errors or even introduce new errors, which makes the detection task more challenging.

\begin{definition}\label{definition: linking patterns}
	\noindent\textbf{Linking Pattern}. For any two triples sharing entities, i.e. $T_1=(h_1, r_1, t_1) \cap T_2=(h_2, r_2, t_2)$, we have two linking patterns: $(i)$ sharing head entity~($h_1 = h_2 \oplus h_1=t_2)$, $(ii)$ sharing tail entity~($t_1 = h_2 \oplus t_1 = t_2$). 
	For our construction criterion, we build two triple graphs with these two linking patterns, based on the rationale that these two patterns possess different semantics.
\end{definition}
In this paper, we propose a novel KG augmentation method that generates different views of KG at triple level and avoids undermining the distribution of potential errors.  Specifically, we create two triple graphs, i.e. $\mathcal{T}$ and $\mathcal{T'}$, by regarding each relational triple as nodes and connect them based on different linking patterns, as shown in  Def~\ref{definition: linking patterns}. These two triple graphs can be regarded as congruent views of the target KG.  Concretely, the triple graph $\mathcal{T}$ models the correlation between relational triples from the perspective of head entity, while the triple graph $\mathcal{T'}$ represent the distribution of relational triple from the perspective of tail entities. Considering that the connected triples that share the same entity are always semantically relevant,
for normal triples in KGs, we can easily find enough relevant neighboring triples in  $\mathcal{T}$, i.e. View I, to reconstruct its semantics learned from $\mathcal{T'}$, i.e. View II. So, modeling and measuring the consistency between its representations learned from these two views can help us assess the trustworthiness of each triple in the original KG.
 \subsection{Error-aware Encoder - EaGNN}
The existence of errors  may significantly jeopardize the expressiveness of the learned representations and consequently degrade the performance of KG error detection, since erroneous triples will also propagate noisy information to its neighbors during the representation learning phase.
Thus, a robust encoder is desired to block the information propagation from potential errors during the message passing process. In this paper, we design a tailored error-aware knowledge graph neural network (EaGNN) as an encoder to capture the rich semantics and structure information of triples in each view. Our EaGNN tightly integrates a local information modeling layer and a global error-aware attention layer that work jointly to get the reliable KG embeddings.
\subsubsection{Local information modeling layer}
Constructing triple graph from the original knowledge graph, in some way, may lose some local structure information, which means a structure of $h \rightarrow r \rightarrow t$. Since every instance in the triple graphs is transformed from a corresponding triple $(h, r, t)$ in the original KG, we first randomly initialize the embedding of entities and relations in the original KG, and then adopt a local information modeling layer, i.e, a set of Bi-LSTM units, to learn the local relational structure within each triple.  Taking the triple $(h, r, t)$ as an example, the local information modeling layer is formulated as follows.
\begin{align}
\label{eq:qi}
    \tilde{\bf e}_h, \tilde{\bf e}_r, \tilde{\bf e}_t & = \text{Bi-LSTM}({\bf e}_h, {\bf e}_r, {\bf e}_t),\\
    {\bf q}_i &= [\tilde{\bf e}_h; \tilde{\bf e}_r; \tilde{\bf e}_t].
\end{align}
The output triple embedding ${\bf q}_i$ could well capture the relational structure within the input triple. Thus, we used them as the initial triple embedding in triple graphs.

\subsubsection{Global error-aware attention layer}
When evaluating the trustworthiness of a triple, apart from the local structures within triples, the global contextual information from the neighbor triples  would also provide crucial hints. 
% The connected triples that share the same entity are always semantically relevant, and thus the degree of acknowledgements from neighboring triples to the target triple reflects whether the target triple can properly integrate into the KG.
% A basic observation is that the anomalous triple is more likely to have fake neighbors that actually are not related to the target triple. 
To model such global contextual information, an advanced approach is to use knowledge graph neural networks (KGNNs). However, existing graph neural networks are not effective for KG error detection because they ignore the existence of anomalous instance in KGs when learning the representations. In other words, potential noisy instances may get the same level of attention as normal ones during the message-passing process, which leads to final triple representations containing noisy facts. 
 To reduce the adverse effect from potential errors, we adopt a novel triple-level attention mechanism to learn global triple representations for each view. 

Given an anchor triple ${\bf q}_{i} \in \mathbb{R}^{d}$ in View I, we update its embedding representation based on the weighted aggregation of its neighbor triples, e.g. $\left\{{\bf q}_{1}, {\bf q}_{2}, \ldots, {\bf q}_{m}\right\}$. The weight between the anchor triple $i$ and its neighbor triple $j$ is calculated as follows.
\begin{equation} \small
\hat{\alpha}_{i j}=\mathcal{A}\left(\mathbf{W} \mathbf{q}_{i}, \mathbf{W} \mathbf{q}_{j}\right).
\label{eq:attention}
\end{equation}
$\hat{\alpha}_{i j}$ indicates the importance of triple $j$ to triple $i$. $\mathbf{W} \in \mathbb{R}^{n \times d}$ is a learnable linear augmentation matrix to project the initial triple representations into the same vector space. $\mathcal{A}$ is the attentional function: $ \mathbb{R}^{n} \times \mathbb{R}^{n} \rightarrow \mathbb{R}$. To make attention coefficients easily comparable across different triples, We normalize them by applying a softmax function.
\begin{equation} \small
\overline{\alpha}_{i j}=\frac{\exp \left(\hat{\alpha}_{i j}\right)}{\sum_{k=1}^{m} \exp \left(\hat{\alpha}_{i k}\right)}.
\label{eq:softmax}
\end{equation}

Since erroneous triples are relatively irrelevant with its neighbors, such an attention mechanism can adaptively assign smaller attention weights to potential abnormal triples. Based on this observation, we introduce a hyperparameter $\mu \in \mathbb{R}$ as threshold value to block information comes from potential anomalous triples. To validate our gated attention mechanism and the effect of $\mu$, we conduct a detailed ablation study in Sec~\ref{sec:abaltion_study} and \ref{sec:parameter}.
\begin{equation} \small
\label{eq:mu}
    \alpha_{i j} ={
    \begin{cases}%\smalll
    \overline{\alpha}_{i j}, & \overline{\alpha}_{i j} > \mu,\\
    0, & \text{otherwise}.
    \end{cases}
    }
\end{equation}

The final error-aware triple representation in View I can be calculated with a sigmoid function, as depicted in Eq.\eqref{eq:xi}:
\begin{equation} \small
\mathbf{x}_{i}=\sigma(\sum\nolimits_{j=1}^{m} \alpha_{i j} \mathbf{W} {\bf q}_{j}).
\label{eq:xi}
\end{equation}

Similarly, given an anchor triple ${\bf q}_{i}$ in View II, we also update its embedding by attending over its neighborhoods' features through:
% Similarly, we could compute the final representation of an anchor triple ${\bf q}_{i}$ in View II through the same encoder.
\begin{equation} \small
\mathbf{z}_{i}=\sigma(\sum\nolimits_{j=1}^{m} \alpha_{i j}' \mathbf{W} {\bf q}_{j}).
\label{final_x}
\end{equation}

\subsection{Joint Confidence Estimation}
To make sure that our model can learn rich and distinguishable across-view representations, we integrate KG embedding loss and standard contrastive learning loss to jointly train the model.

\subsubsection{KG embedding loss}
In triple level, relations can be interpreted as translations operating on the low-dimensional embeddings of the entities. So, the more a triple fits the translation assumption, i.e. $\mathbf{h}+\mathbf{r} \simeq \mathbf{t}$,  the more convincing this triple should be considered.
 Existing KG embedding algorithms have developed various energy functions to model the translational structure for better learning embeddings. In this paper, we take a simple squared euclidean distance to measure the unconformity of each triple with translation assumption and define the KG embedding loss as follows.
\begin{equation} \small
\mathcal{L}_{kge}=\sum_{(h, r, t) \in \mathcal{G}} \sum_{(\hat{h}, \hat{r}, \hat{t}) \in \mathcal{\hat{G}}} \max \left(0, \gamma+E(h, r, t) -E(\hat{h}, \hat{r}, \hat{t})\right),
\label{eq:trans_loss}
\end{equation}
where $E(h, r, t)=\left\|e_{h}+e_{r}-e_{t}\right\|_{2}$ is the traditional energy score for translational embedding models following translation assumption. 
$\gamma > 0$  is the hyperparameter of margin, and $\mathcal{G}$ represents the sampled positive triple set.
Since there are no explicit negative triples in KGs, we generate them by replacing the head $h$ or tail $t$ entity of a triple with a random one $\hat{h}$ or $\hat{t}$.  
\begin{equation} \small
\mathcal{\hat{G}}=\{(\hat{h}, r, t) \mid \hat{h} \in \mathcal{G}\} \cup\left\{\left(h, r, \hat{t}\right) \mid \hat{t} \in \mathcal{G}\right\}.
\end{equation}

\subsubsection{Contrastive loss}
The negative sampling used in KG embedding loss can guide the model to learn rich structure and semantics information within triples. But such kind of local feature alone is not enough to enable effective error detection on the whole KG. To fill the gap, we jointly optimize it along with the contrastive learning loss. By learning and contrasting  between the two triple graphs, our model can learn a rich representation of key semantics in normal instances then the absence of such feature in anomalous triples can force them to lie far in the latent vector space. Therefore, how well a triple’s representations from different views can be pulled close to each other serves as a good learning signal for error detection.

To integrate contrastive learning into KG embedding, in this paper, we employ the normalized temperature-scaled cross entropy loss as our contrastive learning objective to jointly train the model.
% . The core idea is to train the encoder by maximizing the mutual information between two views. 
Specifically, we randomly sample a minibatch of $n$ examples from the original KG and define the training task on pairs of augmented examples derived from the minibatch, resulting in $2n$ triple representations, i.e. $\{{\bf x}_1, {\bf x}_2, ..., {\bf x}_n\}$   from View I and $\{{\bf z}_1, {\bf z}_2, ..., {\bf z}_n\}$ from View II. Assuming that $({\bf x}_i, {\bf z}_i)$ are the representations of triple $i$ learned from different views.
% Instead of sampling negative examples explicitly, given the anchor triple, e.g. ${\bf x}_i$, in View I, we treat ${\bf z}_i$ as positive sample and the other $n-1$ augmented examples in View II as negatives. 
The loss function is defined as follows:
\begin{equation} \small
\mathcal{L}_{con}({\bf x}_i, {\bf z}_i)=-\log \frac{\exp \left(\operatorname{sim}\left({\bf x}_i, {\bf z}_i\right) / \tau\right)}{\sum_{j\in{\{1,2, \dots , n\} \setminus \{i\} }} \exp \left(\operatorname{sim}\left({\bf x}_i, {\bf z}_j\right) / \tau\right)},
\label{con_loss}
\end{equation}
where $\operatorname{sim}({\bf x}_i, {\bf z}_i)$ denotes the cosine similarity of multi-view triple representations, i.e.  ${\bf x}_i$ and ${\bf z}_i$. The final contrastive loss is computed across all triples in a mini-batch.

\begin{algorithm}[t]
%\small
  \caption{Contrastive KG Error Detection}  
    \label{algorithm1}
  \KwIn{Knowledge graph $\mathcal{G}$ with errors.}  
  \KwOut{KG embeddings and triple's confidence score.} 
%   \tcc{Triple representation learning with confidence:} 
%   /* Training the triple embeddings.         */\\
  Initialize network parameters\;  
  Construct two triple graphs $\mathcal{T}$\ and $\mathcal{T'}$\ based on augmentation rules, as shown in Sec~\ref{sec:KG_augmentation}\;
  
  \While{\text{not converged} }
  {\For{$(h,r,t)\in\mathcal{G}$}
  {
    \tcc{Local  information  modeling  layer:}
  Compute the local triple embedding representations, defined in Eqs.\eqref{eq:qi}-(2)\;
  \tcc{Global error-aware attention layer:}
Get global error-aware triple representations, defined in Eqs.\eqref{eq:attention}-\eqref{final_x}\;
\tcc{Joint optimization:}
Jointly optimize the model via a combination of translation-based KG embedding loss in Eq.\eqref{eq:trans_loss} and contrastive loss in Eq.\eqref{con_loss}\;
  }
  }
  \tcc{confidence estimation module:}
Calculate the confidence score of each triple $(h,r, t)$ as defined in Eq. \eqref{confidence_score}\;
\end{algorithm}
\subsubsection{Triple confidence score assessment}
Upon the features learned by our proposed framework, we define the final confidence score function as follows:
\begin{equation}\small
C(h, r, t)= \sigma(\text{sim}({\bf x}_i, {\bf z}_i) - \lambda \cdot E(h, r, t)),
\label{confidence_score}
\end{equation}
    where $E(h, r, t)=\left\|{\bf e}_h+{\bf e}_r-{\bf e}_t\right\|_{2}$ reflects the degree of self contradictory within the triple, while $sim({\bf x}_i, {\bf z}_i)$ measures the consistency of triple representations of the same sample from View I and View II. In practice, we adopt a trade-off parameter $\lambda$ to balance  the  contribution  of these two learning signals for error detection and then a sigmoid function is used to map the score into the range of [0, 1].
    The learning process of CAGED is summarized in Algorithm~\ref{algorithm1}.

\begin{table}[t]
  \caption{The statistical information of the datasets.}
  \label{tab:DatasetStatistics}
\centering
\vspace{-4mm}
\setlength{\tabcolsep}{4.5pt}
% \resizebox{230pt}{30pt}{
\begin{tabular}{lcccc}
\toprule
     Dataset &Entities   &Relations  &Triples    &Mean in-degree \\ \midrule
FB15K &$14$,$541$  &$237$    &$310$,$116$  &$18.71$     \\ 
 WN18RR  &$40$,$943$  &$11$    &$93$,$003$   &$2.12$     \\ 
 NELL-995 &$75$,$492$  &$200$    &$154$,$213$  &$1.98$     \\ \bottomrule
\end{tabular}
% }
\vspace{-4mm}
\end{table}

\renewcommand{\arraystretch}{1} %0.894
\begin{table*}[t]
\centering%\smalll
% \vspace{2pt}
\caption{Error detection results of Precision@K and Recall@K based on the three datasets with anomaly ratio $= 5\%$.}
\setlength{\tabcolsep}{3.3pt}
\vspace{-0.2cm}
%\resizebox{490pt}{100pt}{
\begin{tabular}{llccccccccccccccc}
\toprule
& &\multicolumn{5}{c}{FB15K} &\multicolumn{5}{c}{WN18RR} &\multicolumn{5}{c}{NELL-995} \cr \cmidrule(lr){3-7} \cmidrule(lr){8-12} \cmidrule(lr){13-17} 
& &$K$=$1\%$  &$K$=$2$\%  &$K$=$3$\%  &$K$=$4$\%  &$K$=$5$\%$^a$ & $K$=$1\%$  &$K$=$2$\%  &$K$=$3$\%  &$K$=$4$\%  &$K$=$5$\%$^a$ & $K$=$1\%$  &$K$=$2$\%  &$K$=$3$\%  &$K$=$4$\%  &$K$=$5$\%$^a$ \\ \cmidrule(lr){3-7} \cmidrule(lr){8-12} \cmidrule(lr){13-17} 
% \multirow{9}{*}{$Precision@K$}
\parbox[t]{5mm}{\multirow{7}{*}{\rotatebox[origin=c]{90}{$Precision@K$}}}
& TransE
 &0.756 &0.674 &0.605 &0.546 &0.488 
 &0.581 &0.488 &0.371 &0.345 &0.331
 &0.659 &0.550 &0.476 &0.423 &0.383 \\ 
& ComplEx
 &0.718 &0.651 &0.590 &0.534 &0.485 
 &0.518 &0.444 &0.382 &0.341 &0.307
 &0.627 &0.538 &0.472 &0.427 &0.378  \\ 
& DistMult
 &0.709 &0.646 &0.582 &0.529 &0.483
 &0.574 &0.451 &0.390 &0.349 &0.322
 &0.630 &0.553 &0.493 &0.446 &0.408 \\
& CKRL
 &0.789 &0.736 &0.684 &\underline{0.630} &0.574
 &0.675 &0.526 &0.436 &0.389 &0.349 
 &0.735 &0.642 &0.559 &0.498 &0.450 \\
& KGTtm
 &0.815 &\underline{0.767} &\underline{0.713} &0.612 &\underline{0.579} 
 &\underline{0.770} &\underline{0.628} &\underline{0.516} &\underline{0.444} &\underline{0.396}
 &\underline{0.808} &\underline{0.691} &\underline{0.602} &\underline{0.535} &0.481 \\
& KGIst
 &\underline{0.825} &0.754 &0.703 &0.617 &0.569
 &0.747 &0.599 &0.476 &0.407 &0.379
 &0.782 &0.678 &0.584 &0.528 &\underline{0.485} \\
& CAGED 
 &\textbf{0.852} &\textbf{0.796} &\textbf{0.735} &\textbf{0.665} &\textbf{0.595} 
 &\textbf{0.826} &\textbf{0.726} &\textbf{0.632} &\textbf{0.541} &\textbf{0.469} 
 &\textbf{0.850} &\textbf{0.736} &\textbf{0.644} &\textbf{0.573} &\textbf{0.516} \\ \cmidrule(lr){3-7} \cmidrule(lr){8-12} \cmidrule(lr){13-17} 
\parbox[t]{5mm}{\multirow{7}{*}{\rotatebox[origin=c]{90}{$Recall@K$}}}
&TransE%~\cite{bordes2013translating}
 &0.151 &0.270 &0.363 &0.437 &0.488 
&0.116 &0.195 &0.223 &0.276  &0.331 
&0.132 &0.220 &0.285 &0.338  &0.383  \\
&ComplEx%~\cite{trouillon2016complex} 
&0.143 &0.260 &0.354 &0.427 &0.485 
&0.103 &0.177 &0.229 &0.273 &0.307  
&0.125 &0.215 &0.283 &0.341 &0.378 \\
&DistMult%~\cite{yang2015embedding} 
&0.141 &0.258 &0.349 &0.423 &0.483 
&0.114 &0.180 &0.234 &0.279 &0.322  
&0.126 &0.221 &0.295 &0.357 &0.408 \\
&CKRL%~\cite{xie2018does}  
&0.158 &0.294 &0.410 &\underline{0.504} &0.574 
&0.135 &0.210 &0.261 &0.311 &0.349  
&0.147 &0.256 &0.335 &0.398 &0.450 \\ 
&KGTtm%~\cite{jia2019triple} 
&0.163 &\underline{0.307} &\underline{0.428} &0.490 &\underline{0.579} 
&\underline{0.154} &\underline{0.251} &\underline{0.309} &\underline{0.355} &\underline{0.396}  
&\underline{0.161} &\underline{0.276} &\underline{0.361} &\underline{0.428} &0.481 \\
&KGIst%~\cite{BelthZVK20}
 &\underline{0.165} &0.302 &0.422 &0.494 &0.569
 &0.149 &0.240 &0.285 &0.325 &0.379
 &0.156 &0.271 &0.350 &0.422 &\underline{0.485} \\
&CAGED 
&\textbf{0.171} &\textbf{0.318} &\textbf{0.441} &\textbf{0.532} &\textbf{0.595} 
&\textbf{0.165} &\textbf{0.290} &\textbf{0.379} &\textbf{0.433} &\textbf{0.469} 
&\textbf{0.170} &\textbf{0.294} &\textbf{0.386} &\textbf{0.458} &\textbf{0.516} \\\bottomrule
\end{tabular}\\
% \tabnote{$^{\rm a}$This footnote shows what footnote symbols to use.}
\footnotesize{$^a$ Please note that according to Eq.\eqref{equ:metric}, when $K$ equals the anomaly ratio (i.e. $5\%$), $Precision@K$ and $Recall@K$ are equal.}
% \footnotesize{$^a$ It is noted that results at $K=5\%$ share the same $Precision@K$ and $Recall@K$ since the percentage of errors is also $5\%$. }\\
%}
\label{tab:performance1}
\vspace{-2mm}
\end{table*}

\section{Experiments}\label{sec:exp}
To verify the effectiveness of the proposed framework CAGED, in this section, we conduct comprehensive experiments on three real-world KGs. Specifically, we aim to answer the following questions.
\begin{itemize}
    \item \textbf{Q1 (Effectiveness):} How effective is CAGED compared with the state-of-the-art KG error detection methods?
    \item \textbf{Q2 (Ablation study):} How does each component of CAGED contribute to its performance?
    \item \textbf{Q3 (Parameter analysis):} How do the hyperparameters influence the performance of CAGED?
    \item \textbf{Q4 (Case study):} Is our proposed CAGED able to detect errors in real-world domain-specific KGs?
\end{itemize}

\subsection{Experimental Settings}
In this section, we introduce the detailed experimental settings, including the benchmark datasets, compared baseline methods and evaluation metrics.

\subsubsection{Datasets}
Following the previous study~\cite{xie2018does, jia2019triple}, we employ three real-world datasets that are constructed with noisy triples to be $5 \%, 10 \%$ and $15 \%$ of the whole KGs based on the popular benchmarks, i.e. FB15k, WN18RR and NELL-995. Their statistical information is summarized in Table~\ref{tab:DatasetStatistics}.

\noindent{\bf FB15K} is a dataset adapted from Freebase Knowledge Base. It is expanded with textual relations, which are converted from textual mentions of ClueWeb12 corpus and annotations in Freebase.
% ~\cite{Bollacker-etal07Freebase}

\noindent{\bf WN18RR} is a subset of WordNet, making up for the weakness of WN18. To avoid test leakage, it also excludes inverse relations using the same procedure as the derivation of FB15K. It contains 11 types of relations, and is less complex.
% ~\cite{Heindorf-etal16Vandalism}

\noindent{\bf NELL-995} is derived from the 995th iteration of NELL, which is a huge and growing knowledge base. When constructing this dataset, the triples without reasoning values are eliminated before the Top-200 unique relations are selected. 
% ~\cite{Carlson-etal10Toward}

\subsubsection{Baseline methods}
 In the experiments, we include two categories of baselines. First, KG-embedding based methods, including {\bf TransE}~\cite{bordes2013translating}, {\bf DistMult}~\cite{yang2015embedding}, {\bf ComplEx}~\cite{trouillon2016complex}. To perform error detection, after we have learned the embedding representations, we access the confidence of triples based on the corresponding score functions. For example, in TransE, we use $\|{\bm e}_h+{\bm e}_r - {\bm e}_t\|_2$. Second, state-of-the-art KG error detection methods, i.e., {\bf CKRL}~\cite{xie2018does}, {\bf KGTtm}~\cite{jia2019triple}, and {\bf KGIst}~\cite{BelthZVK20}. {\bf CKRL} advances TransE by taking all paths between the head entity and tail entity into consideration. {\bf KGTtm} further enhances CKRL by integrating the global graph structure of the KG. {\bf KGIst} proposes an unsupervised way to learn soft rules and detect errors based on these rules. 

\subsubsection{Evaluation metrics}
We use ranking measures to evaluate the performance of all the compared approaches. Specifically, we rank all the triples in the target KG according to their confidence score in ascending order. A triple with a lower score would be more likely to be erroneous.  To fairly evaluate the performance of the KG validation, we use the following two evaluation measures.
 
\noindent{\bf Precision@K} represents the percentage of false triples found among the triples with top K lowest confidence scores.
%evaluates the proportion of true errors that we discovered in the top K triples.

\noindent{\bf Recall@K} denotes the percentage of identified false triples over the total erroneous triples.
%measures the proportion of true errors that we discovered in the total number of ground-truth errors.
\begin{equation}\small \label{equ:metric}
    \begin{split}
        \text { $Precision@K$ }&=\frac{\mid \text { Errors Discovered in Top $K$ Ranking List} \mid}{\mid \text { $K$} \mid} \\
    \text { $Recall@K$ }&=\frac{\mid \text { Errors Discovered in Top $K$ Ranking List} \mid}{\mid \text { Total Number of Errors in KG} \mid}
    \end{split}
\end{equation}

\subsubsection{Implementation details}
We implement all the baselines and the proposed framework with PyTorch. For the baseline methods, we use the released codes to conduct experiments. We use Nvidia RTX 3090 GPU server to train our proposed framework as well as all the baselines. To be more specific, we optimize all models with Adam optimizer, where the batch size is fixed at $256$. We use the default Xavier initializer to initialize our model parameters, and the initial learning rate is $0.01$. The embedding size is fixed to $100$ for all models. We apply a grid search for hyperparameter tuning,
% the learning rate is searched from $\{0.05, 0.01, 0.005, 0.001\}$,  
the attention threshold parameter $\mu$ is searched from $0.001$ to $0.2$, the margin parameter $\gamma$ from $0$ to $1$, and the trade-off coefficient $\lambda$ between $0.001$ to $1000$.
The number of neighbors is computed by taking the average number of neighbors of all triples in a dataset to make sure our model could adapt to datasets with different densities of neighbors, and thus make the best of the neighborhood information. 
To reduce the randomness, we  use the random seed and report the average results of ten runs.

\renewcommand{\arraystretch}{1} %0.894
\begin{table*}[t]
\centering%\small
% \vspace{2pt}
\caption{Error detection results on NELL-995 with different anomaly ratios.} \label{tab:ratio}
\setlength{\tabcolsep}{8.5pt}
\vspace{-0.2cm}
\begin{tabular}{llccccccccc}
\toprule
&Ratio &\multicolumn{3}{c}{5\%} &\multicolumn{3}{c}{10\%} &\multicolumn{3}{c}{15\%} \cr \cmidrule(lr){3-5} \cmidrule(lr){6-8} \cmidrule(lr){9-11} 
&$K$ &$K$=$5$\%$^a$  &$K$=$10$\%  &$K$=$15$\% &$K$=$5$\%  &$K$=$10$\%$^a$  &$K$=$15$\%  &$K$=$5$\%  &$K$=$10$\%  &$K$=$15$\%$^a$ \\\cmidrule(lr){3-5} \cmidrule(lr){6-8} \cmidrule(lr){9-11} 
% \multirow{7}{*}{$Precision@K$}
\parbox[t]{5mm}{\multirow{7}{*}{\rotatebox[origin=c]{90}{$Precision@K$}}}
&TransE
 &0.383 &0.285 &0.225
 &0.626 &0.499 &0.407
 &0.702 &0.621 &0.535 \\ 
&ComplEx
 &0.378 &0.289 &0.231
 &0.614 &0.507 &0.402
 &0.696 &0.589 &0.528  \\ 
&DistMult
 &0.408 &0.298 &0.227
 &0.633 &0.510 &0.414
 &0.718 &0.618 &0.548\\
&CKRL
 &0.450 &0.306 &0.236
 &0.679 &0.524 &0.421
 &0.745 &0.646 &0.560\\
&KGTtm
 &0.481 &\underline{0.320} &0.242
 &0.713 &0.527 &0.437
 &0.788 &\underline{0.673} &\underline{0.576}\\
 &KGIst
 &\underline{0.485} &0.317 &\underline{0.244}
 &\underline{0.748} &\underline{0.552} &\underline{0.440}
 &\underline{0.791} &0.663 &0.569\\
&CAGED 
 &\textbf{0.516} &\textbf{0.325} &\textbf{0.251}
 &\textbf{0.799} &\textbf{0.585} &\textbf{0.458}
 &\textbf{0.823} &\textbf{0.729} &\textbf{0.599}\\ \cmidrule(lr){3-5} \cmidrule(lr){6-8} \cmidrule(lr){9-11} 
%\multicolumn{10}{c}{Recall@K}\\ \midrule
%Ratio &\multicolumn{3}{c}{5\%} &\multicolumn{3}{c}{10\%} &\multicolumn{3}{c}{15\%} \cr \cmidrule(lr){2-4} \cmidrule(lr){5-7} \cmidrule(lr){8-10} 
%$K$ &$K$=$5$\%  &$K$=$10$\%  &$K$=$15$\% &$K$=$5$\%  &$K$=$10$\%  &$K$=$15$\%  &$K$=$5$\%  &$K$=$10$\%  &$K$=$15$\% \\ \cmidrule(lr){2-4} \cmidrule(lr){5-7} \cmidrule(lr){8-10} 
% \multirow{7}{*}{$Recall@K$}
\parbox[t]{5mm}{\multirow{7}{*}{\rotatebox[origin=c]{90}{$Recall@K$}}}
&TransE
&0.383 &0.57 & 0.675 
&0.313 &0.499 &0.612 
&0.234 &0.414 &0.535 \\ 
&ComplEx
&0.378 &0.578 &0.693 
&0.307 &0.507 &0.603 
&0.232 &0.393 &0.528  \\ 
&DistMult
&0.408 &0.596 &0.681 
&0.317 &0.510 &0.621 
&0.239 &0.412 &0.548\\
&CKRL
&0.450 &0.612 &0.708 
&0.340 &0.524 &0.632 
&0.248 &0.431 &0.560\\
&KGTtm
&0.481 &\underline{0.640} &0.726 
&0.357 &0.527 &0.656 
&0.263 &\underline{0.449} &\underline{0.576}\\
&KGIst
&\underline{0.485} &0.634 &\underline{0.732}
&\underline{0.374} &\underline{0.552} &\underline{0.660} 
&\underline{0.264} &0.442 &0.569\\
&CAGED 
 &\textbf{0.516} &\textbf{0.650} &\textbf{0.753}
 &\textbf{0.400} &\textbf{0.585} &\textbf{0.687}
 &\textbf{0.274} &\textbf{0.486} &\textbf{0.599}\\ \bottomrule
 
\end{tabular}\\
\footnotesize{$^a$ It is noted that when $K$ equals anomaly ratio, $Precision@K$ and $Recall@K$ both have the same value.}
% \footnotesize{$^a$  It is worth noting that when $K$ equals anomaly ratio, $Precision@K$ and $Recall@K$ both have the same value.}\\
% \footnotesize{$^b$ The smallest spatial unit is county}\\
% \footnotesize{$^c$ The smallest spatial unit is county}\\
%}
\label{tab:performance}
\vspace{-2mm}
\end{table*}

\subsection{Effectiveness of CAGED (\textbf{Q1})}
To answer {\bf Q1}, we conduct comprehensive experiments on three real-world KGs. The experimental results with anomaly ratio equals to $5\%$ are summarized in Table~\ref{tab:performance1}. The results with other anomaly ratios are similar as shown in Table~\ref{tab:performance}. 

In summary, we have three observations. First, KG error detection methods, i.e., CAGED, CKRL, KGTtm, and KGIst are superior to KG-embedding based methods, i.e., TransE, ComplEx, and DistMult. The reason is that these KG embedding frameworks do not consider the errors in KG, thus can not learn discriminative representations for normal and noisy triples.  Second, CAGED achieves the best performance compared with all baselines on the three datasets in terms of recall and precision. For example, comparing with the second best results, CAGED achieves $1.6\%$, $7.3\%$ and $3.1\%$ improvements in FB15K, WN18RR, and NELL-995, respectively, when $K$ equals $5\%$ and anomaly ratio equals $5\%$. It is because all baselines use synthetic false triples to train the model, and thus is hard to detect nontrivial errors with complex patterns.  While our CAGED introduces the multi-view contrasting to perform optimization. So, it is capable of exploiting the relational structure within the triples and cross-view features for error detection. Third, CAGED outperforms all baselines more significantly when the $K$ is smaller.
 To further verify the capability of CAGED in detecting errors in KGs, we conduct a supplementary experiment on NELL-995 with noisy triples to be different ratios of 5\%, 10\% and 15\%. As shown in Table~\ref{tab:ratio}, our model consistently achieves the best performance in all cases, which demonstrates the capability and stability of our model in detecting errors under different scenarios.
\subsection{Ablation Study (\textbf{Q2})} \label{sec:abaltion_study}
% % \vspace{-1cm}
% \subsection{How does each component of CAGED contribute to the performance in error detection.}
We now investigate the question {\bf Q2}. CAGED has three components, i.e., KG augmentation, error-aware encoder (EaGNN), joint estimation. Four pairs of variants of CAGED are used for this ablation study, as shown in Table~\ref{tab:variantsname}. Their detection performance on NELL-995 is included in Table~\ref{tab:variants}. We omit the results on FB15K and WN18RR since they show similar trends.
\begin{table}[t]
%\smalll
  \caption{Four pairs of variants of CAGED.}
  \label{tab:variantsname}
\centering
\vspace{-0.2cm}
\setlength{\tabcolsep}{1.2mm}{}{
% \resizebox{200pt}{42pt}{
\begin{tabular}{lcc} 
\toprule
 & Original component & Replacement\\ \midrule
Var\_DN &    KG augmentation  & Node dropping \\ 
Var\_EP &  KG augmentation  & Edge perturbation \\  \cmidrule(lr){2-2} \cmidrule(lr){3-3}
Var\_Concat & Bi-LSTM units &  Only concatenation\\
Var\_LSTM &  Bi-LSTM units &  LSTM units  \\ \cmidrule(lr){2-2} \cmidrule(lr){3-3}
Var\_GCN & EaGNN & R-GCN\\ 
Var\_GAT & EaGNN & KGAT\\  \cmidrule(lr){2-2} \cmidrule(lr){3-3}
Var\_Local &Joint optimization& Only negative sampling\\
Var\_Global&Joint optimization& Only contrastive learning\\
\bottomrule
\end{tabular}
\vspace{-0.5cm}
}
\end{table}

\subsubsection{Analysis of component 1 - tailored KG augmentation} \label{ablation1}
To demonstrate the effectiveness of the KG augmentation method, we compare it with two other general augmentation approaches, i.e., node dropping and edge perturbation. The corresponding two variants are Var\_DN and Var\_EP. As shown in Table~\ref{tab:variants}, Var\_DN and Var\_EP perform worse, even comparing to some embedding-based baselines, such as TransE. It is because directly applying  node dropping and edge perturbation will destroy the triple structure and introduce noise to KGs, and thus influence training a correct error detector. The huge gap of performance between these two variants and our proposed model supports the assumption that our tailor augmentation mechanism could generate different views on the premise of avoiding undermining the distribution of potential errors, which enables our contrastive framework to perform effective error detection on KGs.

\captionsetup[figure]{}
\begin{figure*}[tbp]
\centering
  \setlength{\abovecaptionskip}{1pt}
  \subfigcapskip=-3pt
  \subfigure[Impact of hyper-parameter $\mu$]{
    \label{fig:mu}
    \includegraphics[width=0.3\textwidth]{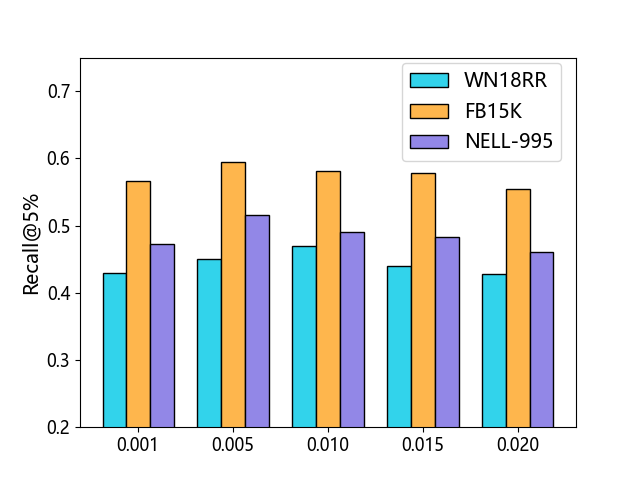}}
  \hspace{2mm}
  \subfigure[Impact of hyper-parameter $\lambda$]{
    \label{fig:lambda}
    \includegraphics[width=0.3\textwidth]{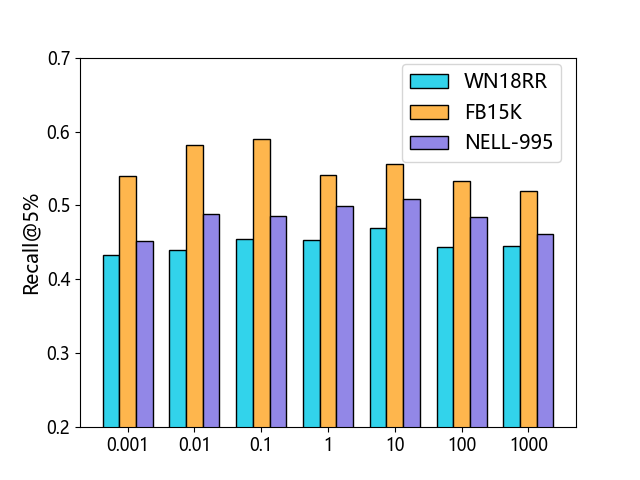}}
  \hspace{2mm}
  \subfigure[Impact of hyper-parameter $\gamma$]{
    \label{fig:gamma}
    \includegraphics[width=0.3\textwidth]{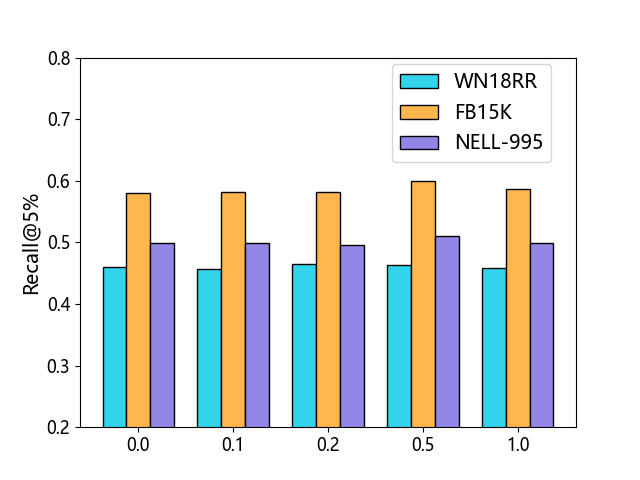}}
    %\vspace{-0.2cm}
  \caption{Impact of hyperparameters on the three datasets.}
\label{fig:parameter}
\vspace{-0.3cm}
\end{figure*}

\subsubsection{Analysis of component 2 - EaGNN} \label{sec:as_EAGNN}
To validate our encoder EaGNN, we replace it with SOTA KGNNs, i.e., R-GCN~\cite{schlichtkrull2018modeling} and KGAT~\cite{nathani-etal-2019-learning}. The corresponding variants are Var\_GCN and Var\_GAT. From Table~\ref{tab:variants}, we can see that the Var\_GAT shows superior performance compared to Var\_GCN. But there is still a significant gap in comparison with CAGED.  It is because that R-GCN is a typical message-passing model designed for KG representation learning. It assumes that all observed triples are correct, and may overfit noisy triples and fail to detect such errors. KGAT adopts an attention mechanism to learn KG representations which may filter out part of noisy information. Nevertheless, Var\_GAT still shows no excellent performance on error detection task. It is because that  KGAT applies attention mechanism from the perspective of entities and relations, while KG errors always occur at the triple level, i.e., the entities naturally exist in KGs, the error is usually a mismatch of the head entity, tail entity, and the corresponding relation. Our tailored  error-aware attention layer in EaGNN takes the triple-level embeddings as input can effectively  filter out noisy triples. 

To evaluate the effectiveness of EaGNN in capturing translation structure within each triple, we introduce Var\_LSTM and Var\_Concat. Var\_LSTM replaces Bi-LSTM units in our local information modeling layers by LSTM. Var\_Concat removes the Bi-LSTM units, and directly concatenates the randomly initialized embeddings of head entity, relation, and tail entity. From the results in Table~\ref{tab:variants}, we observe that the performance of Var\_LSTM and Var\_Concat drops significantly while Var\_LSTM shows better performance compared to Var\_Concat. It is because concatenation ignores the local structure information.

\subsubsection{Analysis of component 3 - joint estimation}
To evaluate the effectiveness of our joint optimization approach, we only use translation based KG embedding loss, i.e., negative sampling,  and contrastive learning loss to train the model in Var\_Local and Var\_Global, correspondingly.  From Table~\ref{tab:variants}, we can observe that both these two variants are inferior to our proposed model while Var\_Local performs worse, even comparing to some straightforward baseline methods, such as DistMult. It is because that negative sampling could only help our model learn rich structure and semantics information within triples, and such kind of local feature is not enough to enable effective error detection on the whole KG.  As a complement to the negative sampling, contrastive learning could help the model learn distinguishable across-view representations. With the rich and  distinguishable across-view representations, error detection on KGs is much more effective.

\begin{table}[t]
\centering%\smalll
% \vspace{3pt}
\caption{The comparisons of CAGED and its variants on NELL-995 with ratio of errors equals $5\%$.}
\setlength{\tabcolsep}{0.7mm}{}{
\vspace{-0.2cm}
\begin{tabular}{lcccccccccc}
\toprule
&\multicolumn{5}{c}{$Precision@K$} &\multicolumn{5}{c}{$Recall@K$}  \\ \cmidrule(lr){2-6} \cmidrule(lr){7-11}
Top@K  &1\%  &2\%  &3\%  &4\%  &5\% &1\%  &2\%  &3\%  &4\%  &5\%\\  \cmidrule(lr){2-6} \cmidrule(lr){7-11}
CAGED 
&\textbf{$.850$} &\textbf{$.736$} &\textbf{$.644$} &\textbf{$.573$} &\textbf{$.516$} 
&$.170$ &$.294$ &$.386$ &$.458$ &$.516$\\  
\cmidrule(lr){2-6} \cmidrule(lr){7-11}
Var\_DN
&$.613$ &$.490$ &$.412$ &$.358$ &$.325$
&$.122$ &$.196$ &$.247$ &$.286$ &$.325$\\
Var\_EP  
&$.593$ &$.500$ &$.416$ &$.354$ &$.313$
&$.118$ &$.200$ &$.249$ &$.283$ &$.313$\\   \cmidrule(lr){2-6} \cmidrule(lr){7-11}
Var\_Concat
&$.767$ &$.648$ &$.532$ &$.468$ &$.440$
&$.153$ &$.259$ &$.319$ &$.374$ &$.440$\\
Var\_LSTM
&$.797$ &$.662$ &$.564$ &$.500$ &$.458$
&$.159$ &$.264$ &$.338$ &$.400$ &$.458$\\ 
\cmidrule(lr){2-6} \cmidrule(lr){7-11}
Var\_GCN 
&$.760$ &$.623$ &$.529$ &$.464$ &$.414$
&$.152$ &$.249$ &$.317$ &$.371$ &$.414$\\ 
Var\_GAT
&$.782$ &$.629$ &$.551$ &$.471$ &$.426$
&$.156$ &$.251$ &$.330$ &$.377$ &$.426$\\  \cmidrule(lr){2-6} \cmidrule(lr){7-11}
Var\_Local
&$.724$ &$.603$ &$.504$ &$.433$ &$.383$
&$.144$ &$.241$ &$.302$ &$.346$ &$.383$\\
Var\_Global
&$.783$ &$.652$ &$.560$ &$.492$ &$.443$
&$.156$ &$.261$ &$.336$ &$.393$ &$.443$\\ 
\bottomrule 
\end{tabular}}
\label{tab:variants}
\vspace{-0.5cm}
\end{table}

\subsection{Parameter Analysis (\textbf{Q3})} \label{sec:parameter}
In this section, we mainly investigate the impact of three key parameters in our framework and report the results in Figure~\ref{fig:parameter}.

As discussed in Eq.\eqref{eq:mu}, $\mu$ is the parameter that controls the attention threshold. The smaller $\mu$ allows more irrelevant neighbors to be attended over when learning the global triple embedding representations, while a larger $\mu$ forces our encoder to focus on a few but closely related neighbor triples. To investigate the impact of $\mu$, we vary it from $0.001$ to $0.020$. Figure~\ref{fig:mu} shows the detection performance in terms of Recall@K as the evaluation metrics. When $\mu = 0.020$, performance on all three datasets is not ideal enough. This happens because only few global contextual features have been mined when $\mu$ is large. On the missing of enough contextual information, the error detection task will be very challenging. As $\mu$ decreases, our encoder starts to attend over more neighbor triples to learn much richer global information, so the performance keeps improving. As shown in Figure~\ref{fig:mu}, performance on FB15K and NELL-995 achieves optimal when $\mu$ is close to $0.005$ while WN18RR gets the best performance at $\mu = 0.01$. However, as $\mu$ continues to fall down, the contrary result happens, i.e., the performance on these three datasets markedly decreases. This is because smaller $\mu$ allows more irrelevant neighbors involving in the representation learning process, correspondingly, some potential noisy triples may swoop in, which leads to overfitting issues for error detection.

\begin{table}[t] %\small
% \vspace{2pt}
  \caption{The average running time for one iteration.}
\centering
\vspace{-0.2cm}
\setlength{\tabcolsep}{9.5pt}
 \begin{tabular}{lccc} 
 \toprule[\heavyrulewidth]
 \textbf{Method} &\textbf{FB15K} &\textbf{WN18RR} &\textbf{NELL-995} \\ \midrule
 TransE &\textbf{0.05} &\textbf{0.07} &\textbf{0.09}\\
 ComplEx &0.06 &0.09 &0.12\\
 DistMult &0.07 &0.11 &0.14\\ \midrule
 CKRL &0.45 &0.29 &0.32\\
 KGTtm &0.44 &0.32 &0.42\\
 KGIst &0.57 &0.49 &0.49\\
 CAGED &0.41 &0.22 &0.38\\ \bottomrule
 \end{tabular}
\label{table:efficiency}
\vspace{-0.5cm}
\end{table}

As shown in Eq.\eqref{confidence_score}, $\lambda$  is the trade-off coefficient that balances the contribution of two learning signals for error detection, i.e., the inconsistency of triple representations across multi-views (e.g., ${\bf x}_A$ and ${\bf z}_A$) and the self-contradictory within the triple embedding $({\bf e}_h, {\bf e}_r, {\bf e}_t)$. The larger $\lambda$ indicates that effective error detection relies more on the signal of the self-contradictory, and vice versa. To investigate the impact of $\lambda$, we vary it from $10^{-3}$ to $10^{3}$ with a multiple 10. We perform detection task on all datasets and the detection performance in terms of Recall@K is shown in Figure~\ref{fig:lambda}. From the results, we observe that 1) WN18RR achieves optimal result when $\lambda$ is around $10$, but, from the global perspective, the performance on WN18RR seems to be not sensitive with parameter $\lambda$ since its value is stable basically without great fluctuate, similar results are observed on NELL-995. 2) For FB15K, the detection performance is much better when the value of $\lambda$ is less than $1$, with an optimal result at $0.1$. After a detailed  analysis of these three datasets, we figure out that the potential reason for this observation is that FB15K has a more dense data structure compared with the other datasets. Specifically, there exists more triple-level neighbors in such a dense data structure  which assistant our model to learn more comprehensive  global triple embedding representations from different views. With richer triple representations, the learning signal of the inconsistency across multi-views can play a more important role in detection task.
% FB15K, WN18RR and NELL-995 achieve optimal results at $\lambda$ is around $0.1, 10, 10$ respectively, and . This indicates that error detection on FB15K

 $\gamma$ serves as a margin to control the distance between the representations of negative and positive pairs in the translation-based KG embedding loss function, i.e., Eq.\eqref{eq:trans_loss}.  To investigate the impact of $\gamma$, we vary it from $0.1$ to $1.0$. Figure~\ref{fig:gamma} shows the detection performance is stable basically without great fluctuation as $\gamma$ varies. This is mainly because we adopt a joint training approach,  avoiding the model from falling into suboptimal solutions.
% 106
% to 103. As shown in Figure~\ref{fig:lambda}, we choose values of $\lambda$ from 0.001 to 1000 with a multiple 10 and set $\mu$ = $0.010$,$\gamma$ = $0.1$. From the results in Figure~\ref{fig:lambda}, we have three major observations.

% \begin{figure}[t!]
% \begin{center}
% \includegraphics[scale=0.42]{resources/Visualization1.png}
% \end{center}
%  \vspace{-0.2cm}
% 	\caption{The scatter plot of the triple confidence values
% distribution.}
% 	\label{fig:augmentation}
% 	\vspace{-0.4cm}
% \end{figure}

\captionsetup[figure]{}
\begin{figure}[tbp]
\centering
  \setlength{\abovecaptionskip}{1pt}
  \subfigcapskip=-5pt
  \subfigure[Visualization of our CAGED]{
    \label{fig:cs1}
    \includegraphics[width=0.21\textwidth]{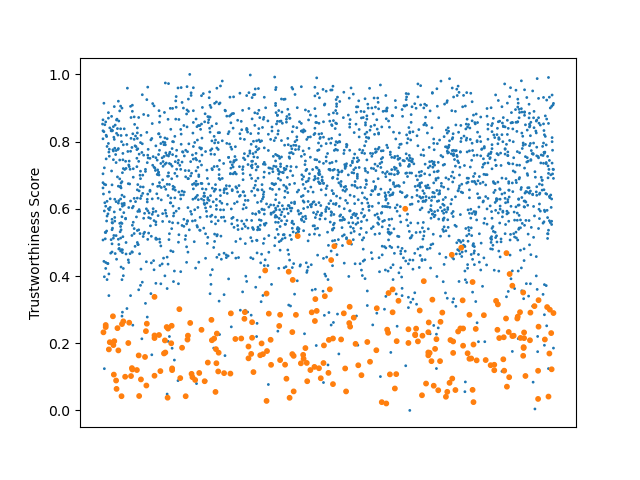}}
  \hspace{2mm}
  \subfigure[Visualization of KGTtm]{
    \label{fig:cs2}
    \includegraphics[width=0.21\textwidth]{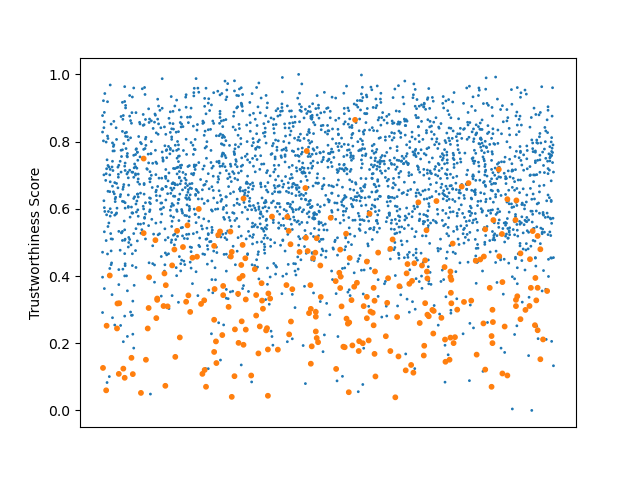}}
  \hspace{2mm}
  \caption{The scatter plot of confidence scores distribution.}
\label{fig:case_study}
\vspace{-4mm}
\end{figure}

\subsection{Case Study (\textbf{Q4})}
The three benchmark datasets used in this paper are all general-purpose KGs.
To further verify the effectiveness of our proposed CAGED, we conduct a case study on another real-world domain-specific knowledge graph, KnowLife~\cite{ernst2015knowlife}. KnowLife is a huge medical knowledge repository, which is automatically constructed based on scientific literature. 
To access ground-truth labels of errors, we manually label triples that are randomly selected from the  official KnowLife website.
% \footnote{\url{http://knowlife.mpi-inf.mpg.de/}}.
Concretely, we first employ two participants to manually label each triple by cross-checking with the information presented on the original website. If the labels  provided by these two participants are inconsistent, then this triple will be assigned to the third one for further investigation. Majority voting is applied to decide the final label of each triple. Following this procedure, we label more than 5000 triples, which are divided into two sets, i.e., the set of positive samples and the set of noisy samples. 

To make the comparison more simple and intuitive, we display the triple effectiveness scores learned by our model and the best baseline method, i.e. KGTtm, in a centralized coordinate system, as shown in Figure~\ref{fig:case_study}. Specifically, the Figure~\ref{fig:cs2} in the right visualizes the distribution of the values of the normal and noisy triples learned by KGTtm. It can be seen that a considerable proportion of erroneous triples have been wrongly classified with high confidence scores. But for our model, as shown in Figure~\ref{fig:cs1}, the confidence value of normal triples are mainly distributed in the upper region, while the values of the erroneous triples are mainly concentrated in the lower region, which verifies the superiority of our CAGED for KG error detection in real-world scenarios.

\subsection{Efficiency Analysis of CAGED}
To measure the efficiency of CAGED, we record the average computation time of one iteration for all models under the same setting, and report the results in Table~\ref{table:efficiency}. All experiments are conducted with one Nvidia RTX $3090$ GPU.
From Table~\ref{table:efficiency}, we can observe that embedding methods including TransE, ComplEx, and DistMult run faster than KG error detection methods in general since they only need to calculate the mean square losses. TransE costs less time than ComplEx and DistMult on all datasets. Comparing to other three error detection methods, our model shows comparable efficiency on average, while runs faster especially in the large-scale dataset, FB15K. This observation validates the efficiency of our model comparing to baseline methods. In summary, our model is more promising for KG error detection in real-world scenarios,  due to the fact that it outperforms state-of-the-art KG error detection algorithms, with comparable or even better running efficiency.

\section{Related Work}
Recent studies in KG error detection could be roughly categorized into two classes. First, {\bf classification-based} models. Melo and Paulheim~\cite{Melo-Paulheim17Detection} design a classifier based on entity categories and features of related paths. Jia et al.~\cite{jia2019triple} integrate features from entity level, relation level, and global level to build a classifier. Ge et al.~\cite{Ge-etal20KGClean} explore using active learning to train a classifier based on a small number of labels. A remaining issue of classification-based models is the penalty of imbalanced classes, because there are far more correct triples than errors. Second, {\bf embedding-based} models. TransE~\cite{bordes2013translating} is a typical KG embedding algorithm. It uses margin loss and negative sampling to induce ${\textbf e}_h + {\textbf e}_{r} \approx {\textbf e}_t$. Several studies~\cite{Guo-etal20Measuring,Zhao-etal19SCEF,Wang-etal2020Efficient} propose to evaluate the confidence of triples based on the embedding loss, e.g., ${\textbf e}_h + {\textbf e}_{r} - {\textbf e}_t$.

Efforts have also been devoted to employing extra information to facilitate KG error detection. In some KGs, categories of entities are (partially) available. Paulheim and Bizer~\cite{paulheim2014improving} employ the statistical distributions of relations and entity categories to assess the suspiciousness of triples. Attempts have been made to leverage entity categories to perform clustering-based outlier detection~\cite{Paulheim-Gangemi15Serving} or construct path-based rules~\cite{shi2016discriminative}. However, entity categories are not available in many KGs. Other beneficial extra information includes related webpages~\cite{Lehmann2012defacto}, crowdsourcing evaluations~\cite{Jeyaraj-etal19Probabilistic}, and external human-curated KGs~\cite{Wang-etal2020Efficient}.

\section{Conclusions and Future Work}\label{sec:con}
Existing KG error detection methods mainly rely on synthetic false triples to train their models, which are generated by replacing the head or tail entity in a triple with a random entity. These synthetic errors are not in line with the ones in practice, e.g., (Bruce\_Lee, place\_of\_birth, Chinatown), in which the three elements in a triple are highly relevant. To this end, we propose an effective framework named CAGED, which integrates contrastive learning and KG error detection. CAGED provides a novel view towards knowledge graph modeling. It creates different hyper-views of the same graph structure by centering triples, and then learns the representation by maximizing agreement of each instance's representation between these views. By learning to contrast elaborate instance pairs, the model can acquire informative features that can be used as an effective learning signal to detect errors in KGs. 
Experimental results on three real-world KGs demonstrates that our model outperforms the state-of-the-art KG error detection algorithms in terms of different evaluation metrics.
In the future, we will develop error-aware KG embedding model to facilitate KG-related downstream tasks on the basis of CAGED.
\begin{acks}
The authors gratefully acknowledge receipt of the following financial support for the research, authorship, and/or publication of this article. This work was supported in full by the Hong Kong Polytechnic University, Departmental General Research Fund (Project Number: P0036738).
\end{acks}

% \newpage
% \section*{Acknowledgment}
% \bibliographystyle{}
\balance
% \bibliographystyle{ACM-Reference-Format}
%   \bibliography{abbrv_reference.bib}
\bibliographystyle{unsrt}
\bibliography{paper.bib}

\end{document}